\documentclass[10pt,twocolumn,letterpaper]{article}

\usepackage[pagenumbers]{iccv}  

\usepackage{multirow} 

\DeclareGraphicsExtensions{.pdf,.png}

\definecolor{iccvblue}{rgb}{0.21,0.49,0.74}
\usepackage[pagebackref,breaklinks,colorlinks,allcolors=iccvblue]{hyperref}

\title{GLIMPSE: Holistic Cross-Modal Explainability for Large Generative Vision–Language Models}

\author{Guanxi Shen\\
Georgia Institute of Technology\\
}

\begin{document}
\maketitle
\begin{abstract}
Recent large vision--language models (LVLMs) has advanced capabilities in visual question answering (VQA). 
However, interpreting where LVLMs direct their visual attention remains a challenge, yet is essential for understanding model behavior. 
We introduce \textbf{GLIMPSE} (Gradient-Layer Importance Mapping for Prompted Visual Saliency Explanation), 
a lightweight, model-agnostic framework that jointly attributes LVLM outputs to the most relevant visual evidence and textual signals that support open-ended generation.
GLIMPSE fuses gradient-weighted attention, adaptive layer propagation, and relevance-weighted token aggregation to produce holistic 
response-level heat maps for interpreting cross-modal reasoning, outperforming prior methods in faithfulness and pushing the state-of-the-art in human-attention alignment. 
We demonstrate an analytic explainable AI (XAI) approach to uncover fine-grained insights into LVLM cross-modal attribution, 
trace reasoning dynamics, analyze systematic misalignment, diagnose hallucination and bias, and ensure transparency.

\begin{figure*}[t]
    \centering
    
    \begin{minipage}{0.97\textwidth}
    \raggedright
    {\footnotesize\textbf{Q:} \textcolor{black}{What} \textcolor{red!83}{color} \textcolor{red!50}{are} \textcolor{black}{the} \textbf{\textcolor{red!85}{shoes}} \textcolor{red!62}{for} \textcolor{red!50}{woman} \textcolor{black}{on} \textcolor{red!67}{left?} \quad \textbf{A:} The \textbf{\textcolor{orange!70}{shoes}} for the woman on \textcolor{orange!55}{left} \textbf{\textcolor{orange!88}{are}} \textbf{\textcolor{orange!88}{yellow.}}}
    \end{minipage}
    
    \vspace{0.3em}
    \includegraphics[width=0.97\textwidth]{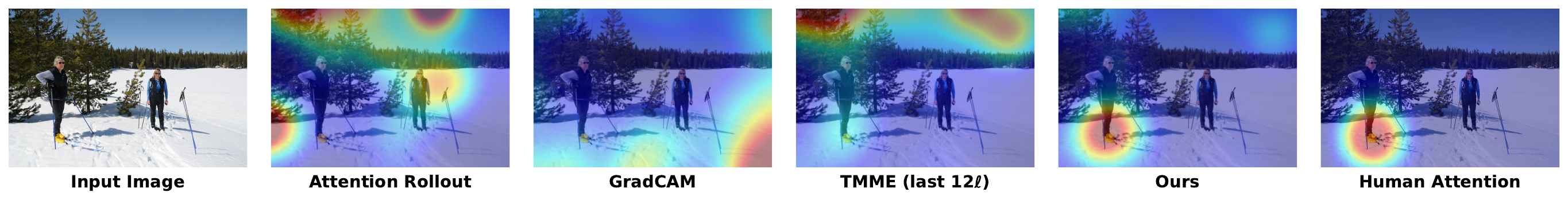}
    
    \vspace{0.4em}
    
    \begin{minipage}{0.97\textwidth}
    \raggedright
    {\footnotesize\textbf{Q:} \textcolor{black}{What} \textbf{\textcolor{red!100}{material}} \textcolor{black}{are} \textcolor{black}{the} \textcolor{black}{man's} \textcolor{red!65}{pants} \textbf{\textcolor{red!96}{made}} \textbf{\textcolor{red!100}{from?}}  \textbf{A:} The man's pants are made from \textbf{\textcolor{orange!98}{denim,}} indicated by \textcolor{orange!60}{blue} color and typical \textbf{\textcolor{orange!80}{texture}} of jeans.}
    \end{minipage}
    
    \vspace{0.3em}
    \includegraphics[width=0.97\textwidth]{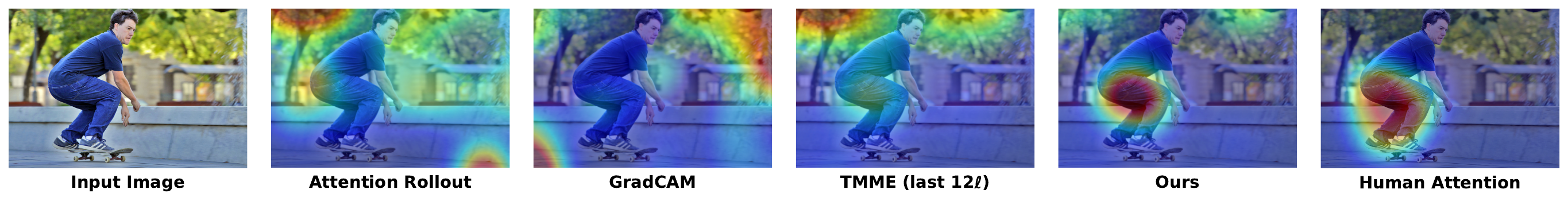}
    
    \vspace{0.4em}
    
    \begin{minipage}{0.97\textwidth}
    \raggedright
    {\footnotesize\textbf{Q:} \textcolor{black}{What} \textbf{\textcolor{red!71}{brand}} \textcolor{black}{of} \textbf{\textcolor{red!59}{laptop}} \textcolor{black}{is} \textcolor{black}{that?} \quad \textbf{A:} The laptop in the image \textcolor{orange!65}{is} an \textbf{\textcolor{orange!75}{Apple}} \textcolor{orange!65}{MacBook,} identifiable by \textbf{\textcolor{orange!95}{logo}} on back of screen.}
    \end{minipage}
    
    \vspace{0.3em}
    \includegraphics[width=0.97\textwidth]{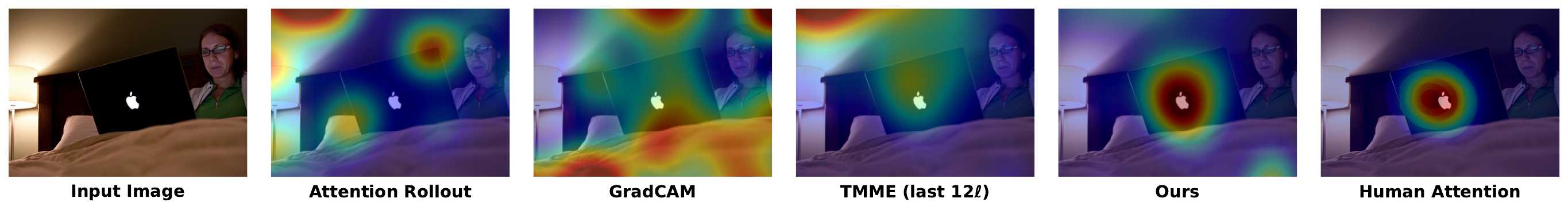}
    
    \vspace{0.4em}
    
    \begin{minipage}{0.97\textwidth}
    \raggedright
    {\footnotesize\textbf{Q:} \textcolor{black}{What} \textbf{\textcolor{red!100}{color}} \textcolor{black}{is} \textcolor{black}{the} \textbf{\textcolor{red!93}{vehicle}} \textcolor{red!66}{in} \textbf{\textcolor{red!100}{mirrored}} \textcolor{red!50}{image?}  \textbf{A:} \textcolor{black}{The} \textcolor{black}{vehicle} \textcolor{black}{in} \textcolor{black}{the} \textcolor{orange!75}{mirrored} \textcolor{orange!52}{image} \textbf{\textcolor{orange!70}{is}} \textbf{\textcolor{orange!99}{white.}}}
    \end{minipage}
    
    \vspace{0.3em}
    \includegraphics[width=0.97\textwidth]{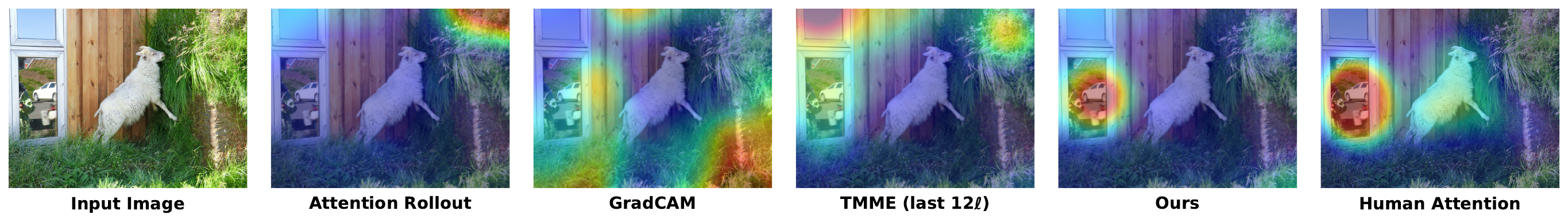}

    \caption{\textbf{Comparison.} Qualitative comparison between our method and baseline explainers on VQA samples. The coloring indicates token relevancy weighting which is only applied to \textbf{Ours} method.}
    \label{fig:teaser}
    \vspace{-0.1em}
\end{figure*}

\end{abstract}
    
\section{Introduction}
\label{sec:intro}

Recent large vision–language models (LVLMs) \cite{liu2023llava,bai2023qwenvl} have demonstrated the 
ability to generate open-ended textual responses based on visual inputs. These systems can cite objects, describe scenes, and follow multi-step reasoning prompts with a level of coherence that was out of reach only a few years ago. Yet the internal reasoning mechanisms that enable such visual–textual capability remain largely opaque.

Interpreting precise visual attribution can expose spurious correlations, reveal bias and hallucinations, and provide insights into understanding model behavior. Human-gaze studies in visual question answering show that models whose learned attention aligns with human fixations—or are explicitly tuned to do so—tend to achieve higher accuracy, suggesting that interpretability and task performance are intertwined \cite{xu2022vqamhug,selvaraju2019hint,qiao2018attention,sood2023mulan}.

A spectrum of explanation techniques has been adapted to multimodal Transformers. Attention-based methods—from raw cross-attention maps to Attention Rollout \cite{abnar2020attention}—are efficient, 
yet often produce noisy, non-causal hotspots. Gradient-based methods such as Gradient × Input \cite{simonyan2014deep}, Grad-CAM \cite{selvaraju2017gradcam} and Integrated Gradients \cite{sundararajan2017axiomatic} inherit noisy gradients when applied to deep architectures and may suffer from low faithfulness. 
Layer-wise Relevance Propagation (LRP) methods \cite{ali2022xai,achtibat2024attention} raise implementation complexity with limited gains in interpretability or accuracy. Perturbation and iterative approaches iGOS++ \cite{khorram2021igos}, SHAP \cite{lundberg2017unified} and PixelSHAP \cite{petsiuk2023pixelshap}
estimate input importance by measuring output change under content removal. These methods typically provide label- or token-specific attributions, and some incur significant computational overhead and scale poorly with sequence length. 
Each method family offers complementary insights, yet they may fall short in providing efficient and scalable global explainability for modern generative LVLMs that produce multi-sentence, autoregressive responses.

To address these challenges, we propose \textbf{GLIMPSE} (Gradient-Layer Importance
Mapping for Prompted Visual Saliency Explanation).  To our knowledge, this is the first
attention–gradient framework capable of explaining entire free-form LVLM responses.

\noindent Our main contributions are:
\begin{enumerate}
    \item We introduce GLIMPSE, a lightweight, model-agnostic, gradient-attention framework that produces a faithful, holistic attribution map for free-form LVLM outputs, explaining both visual and textual saliency.
    \item We establish state-of-the-art human-attention alignment for LVLM interpretability, outperforming prior methods by over 46\% in rank correlation and 71\% in Normalized Scanpath Saliency (NSS) on VQA-HAT (Table~\ref{tab:quantitative_results}).
    \item We demonstrate GLIMPSE’s diagnostic capabilities through case studies that reveal cross-modal reasoning and systematically investigate attention misalignment, hallucinations, and bias.
  \end{enumerate}

\section{Related Works}
\label{sec:related}

\subsection{Attention-Based}
\label{subsec:attention_methods}
Transformers expose an intuitive signal in their attention weights, and early multimodal works therefore projected raw cross-attention maps as saliency. However, these maps are known to explain only a subset of the model's computation and lack a strong causal relationship to the output. 
Attention Rollout \cite{abnar2020attention} propagates the weight matrices of successive layers, improving information flow but at the cost of amplified noise, especially for deeper networks. 

\subsection{Gradient-Based}
\label{subsec:gradient_methods}
Another line of work treated the gradient of the class logit with respect to each visual token as an importance signal \cite{simonyan2014deep}, often visualized as Gradient × Input. 
While conceptually simple, raw gradients fluctuate strongly across layers, a phenomenon later termed gradient shattering \cite{balduzzi2017shattered}, yielding noisy and speckled heatmaps. 
Grad-CAM \cite{selvaraju2017gradcam} alleviates this by weighting the last-layer feature map with the spatially averaged positive gradients, 
producing coarse yet class-aligned localization. 
When applied to multimodal Transformers, however, gradient-based methods often suffer from vanishing or oscillatory signals along deep layers, 
resulting in fragmented and low-faithfulness heatmaps.

\subsection{Propagation-Based}
\label{subsec:propagation_methods}

While Attention Rollout offers a lightweight heuristic by propagating attention multiplicatively, 
Layer-wise Relevance Propagation (LRP) \cite{bach2015pixel} propagates additively and enforces relevance conservation across layers. 
Naive Transformer adaptations of LRP break conservation due to softmax non-linearities, yielding unstable, 
saturated heatmaps. AttnLRP \cite{achtibat2024attention} mitigates this issue by detaching the softmax and propagates relevance only through the value path. 
Despite increased computational demand and implementation complexity, LRP-based techniques may not offer improved attribution, as demonstrated by Chefer et al.~\cite{chefer2021generic}, 
who advocate a more streamlined propagation scheme, Generic Attention-Model Explainability (TMME) in preference. 

TMME represents a hybrid approach, fuses the positive gradient with the attention weights, and additively propagates this relevance through layers, yielding more locally grounded maps with cross-modal relevances. 
Yet, like most Transformer explainers, it was originally designed to compute saliency for a single target, and thus does not inherently provide a unified picture of how visual evidence accumulates across an entire sequence. 
Moreover, when propagated through the much deeper stacks of modern LVLMs, its relevance can fragment and amplify noise, leading to degraded performance (Sec.~\ref{sec:experiments}). 
Nevertheless, TMME’s core premise provides important inspiration, which GLIMPSE extends and enhances for LVLMs.

\subsection{Perturbation-Based}
\label{subsec:perturbation_methods}
Perturbation-based methods explain a model by masking parts of the input and observing the change in its output. SHAP \cite{lundberg2017unified} approximates Shapley values by sampling many masked input subsets. When transferred to multimodal Transformers, these approaches preserve their theoretical faithfulness but incur a steep computational cost: the number of forward passes grows significantly with image resolution and sequence length, rendering them impractical for long, free-form generative outputs.
Perturbation-based hybrid methods including Iterated Integrated Attributions \cite{barkan2023visual} refine Integrated Gradients \cite{sundararajan2017axiomatic} by re-integrating gradients along internal layers. AtMan \cite{deb2023atman} perturbs a Transformer's own attention matrices to derive relevance maps. IGOS++ \cite{khorram2021igos} optimizes a saliency mask with integrated-gradient guidance plus bilateral perturbations. Nonetheless, these hybrid methods also impose significant memory and computational overhead, limiting their practical adoption.

\subsection{Current Explainability Methods for LVLMs}
\label{subsec:current_lvlm_methods}
Explainability for generative LVLMs remains a relatively underexplored area, yet recent methods have been proposed to begin closing the gap. LVLM-Interpret \cite{benmelech2024lvlm} visualizes raw cross-attention maps and gradient relevancy, thus inheriting the well-known non-causality and noisy artifact issues, and furthermore provides only token-level heatmaps. Q-GroundCAM \cite{rajabi2024qgroundcam} applies GradCAM to quantify phrase grounding, offering quick gradient-based maps yet still focusing on token/phrase-level grounding. 
PixelSHAP \cite{petsiuk2023pixelshap} extends SHAP to segmentation masks, producing global saliency maps but remaining computationally intensive. An LVLM-specific IGOS++ variant \cite{xing2025attention} similarly yields a holistic heat-map for each free-form answer, albeit through costly iterative optimisation and offers limited interpretability.
Architectural approaches \cite{giulivi2024owl} embed object detectors into an VLM to generate built-in saliency but at the expense of architectural modifications and additional training.
Collectively, existing explanation methods are either token-/phrase-centric or rely on costly perturbation, and thus fall short of comprehensively addressing the distinct challenges posed by generative LVLMs.

\subsection{Challenges for Interpreting LVLMs}
\label{subsec:lvlm_challenges}
Modern generative LVLMs introduce four key challenges for saliency explanation that go beyond those faced in non-autoregressive or single-output vision–language models:

\textbf{Multi-sentence decoding:} as the model autoregressively emits a free-form answer, its visual focus shifts over time, 
explanations therefore must be aggregated across the entire sequence, rather than individual token level. 

\textbf{Cross-modal token entanglement:} Visual and textual tokens are interleaved, requiring an attribution scheme that simultaneously respects both modalities and interprets their joint importance.

\textbf{Architectural depth:} Deep Transformer stacks amplify noise during naive relevance propagation, producing checkerboard artifacts that obscure causal attributions.

\textbf{Long contexts:} Extended input–output contexts inflate sequence length, making costly perturbation and iterative optimization methods impractical and further diminishing the interpretability of token-level attributions.

These open challenges underscore the need for a lightweight, holistic, sequence-level interpretability framework that respects cross-modal interactions and remains robust to the deep Transformer architectures typical of modern LVLMs—a gap that GLIMPSE is designed to address.
\section{Method}

GLIMPSE operates in three stages.

\begin{enumerate}
\item \textbf{Layer Relevance Extraction:} Within each layer, we weight attention score by its positive gradient, then fuse across heads using weights proportional to head importance, producing a layer-wise relevance map.

\item \textbf{Adaptive Layer Propagation:} These layer relevance maps are propagated through the layers using composite weights factoring each layer’s gradient norm and a depth-based prior.
\item \textbf{Cross-Modal Token Relevancy:} Token relevance is rescaled by prompt alignment, visual grounding, and its softmax confidence, then aggregated across the sequence into a unified response-level saliency map.

\end{enumerate} 

GLIMPSE is model-agnostic and attaches to any autoregressive vision–language model. A full explanation requires one forward pass to generate the response and extract attention tensors, followed by one backward pass per generated token to compute gradients.

\subsection{Preliminaries}

We consider an autoregressive vision--language model that takes a single image $I$ and a textual prompt $p$, then generates a free-form response $y_{1:T}$. The model comprises $L$ Transformer blocks, each with $H$ attention heads.

\paragraph{Sequence representation} The visual tokens $v_{1:K}$ are concatenated with the prompt tokens $p_{1:M}$ and the generated tokens $y_{1:T}$ into one causal sequence
\begin{equation}
\label{eq:sequence_repr}
x = [v_{1:K} \| p_{1:M} \| y_{1:T}],
\end{equation}
whose length is $N = K + M + T$. The index sets are
\begin{align}
\mathcal{V} &= \{1,\ldots,K\} \text{ for image tokens,}\\
\mathcal{P} &= \{K+1,\ldots,K+M\} \text{ for prompt tokens,}\\
\mathcal{Y} &= \{K+M+1,\ldots,N\} \text{ for generated tokens.}
\end{align}

\paragraph{Attention tensors} For layer $\ell$ and head $h$, the attention matrix $A_{\ell}^{h} \in \mathbb{R}^{N \times N}$ stores the softmax-normalized dot-product between queries and keys.

\paragraph{Gradients} We denote by $g_{\ell}^{h} = \frac{\partial z_t}{\partial A_{\ell}^{h}}$ the gradient of the logit $z_t$ corresponding to the target token $t$ with respect to the attention weights of head $h$ in layer $\ell$. 

\paragraph{Goal} From the set of attention maps $\{A_{\ell}^{h}\}$ and their gradients $\{g_{\ell}^{h}\}$, GLIMPSE computes
\begin{enumerate}
\item \textbf{dual modality saliency maps}: visual saliency $\tilde{\mathbf{R}}_\mathcal{V}$ highlighting image regions most responsible for the generation, and prompt saliency $\tilde{\mathbf{R}}_\mathcal{P}$ quantifying how prompt components guide visual attention; and
\item \textbf{cross-modal token relevance scores} $\gamma_t$ for $t \in \mathcal{Y}$ that capture each generated token's joint alignment with both visual content and prompt context;
\end{enumerate}

\subsection{Layer Relevance Extraction}
\label{sec:head_fusion}
Attention heads within each Transformer layer may not contribute uniformly to the model's output. Therefore, we construct a fused relevance map for each layer by integrating local and global weighting signals.

Following Chefer~\emph{et al.}~\cite{chefer2021generic}, for head $h$ in layer $\ell$, we take the element-wise product of its attention matrix $A_{\ell}^{h}$ and the corresponding positive gradient $g_{\ell}^{h}$:

\begin{equation}
\label{eq:gw_pos}
G_{\ell}^{h} \;=\; \operatorname{ReLU}\!\bigl(g_{\ell}^{h}\odot A_{\ell}^{h}\bigr),
\end{equation}
to highlight local positions that both attend strongly and receive a positive contribution from the backward signal.

Instead of uniform head averaging used by \cite{chefer2021generic}, we apply a global head-weighting scheme that emphasizes heads with higher contribution. Each head's contribution is quantified by aggregating its gradient-weighted attention scores and normalizing by the total positive gradient mass:
\begin{equation}
\label{eq:head_weight}
w_{\ell}^{h} = \text{softmax}\left(\frac{1}{\lambda} \cdot \frac{\sum_{i,j} G_{\ell}^{h}(i,j)}{\sum_{i,j} \operatorname{ReLU}\bigl(g_{\ell}^{h}(i,j)\bigr)}\right),
\end{equation}
where $\lambda$ is the temperature for softmax. Observe that
\begin{equation}
\label{eq:head_expectation}
\frac{\sum_{i,j} G_{\ell}^{h}(i,j)}{\sum_{i,j} \operatorname{ReLU}\bigl(g_{\ell}^{h}(i,j)\bigr)}
\;=\;
\mathbb{E}_{(i,j)\sim g_{\ell}^{h+}}\!\bigl[A_{\ell}^{h}(i,j)\bigr],
\end{equation}
where the expectation is over positions $(i,j)$ weighted by the positive gradients $g_{\ell}^{h+} = \operatorname{ReLU}\bigl(g_{\ell}^{h}\bigr)$. This ratio represents the expectation of the head's attention under the positive-gradient distribution, hence is large only when the head concentrates attention on gradient-relevant positions.
Globally, this weight measures which heads have the strongest overall positive-gradient support.

The fused attention matrix for layer $\ell$ is computed as:
\begin{equation}
\label{eq:fused_attention}
E_\ell = \sum_{h=1}^{H} w_{\ell}^{h} G_{\ell}^{h},
\end{equation}
which is then row-normalized to preserve probability mass.

\subsection{Weighted layer propagation}
\label{sec:layer_weight}

\paragraph{Adaptive layer weighting}
To propagate relevance across layers, we introduce a combined weighting that considers both gradient magnitude and layer depth. We define
\begin{equation}\label{eq:layer_gradient_norm}
g_\ell = \left\|\sum_{h=1}^{H} g_{\ell}^{h}\right\|_1
\end{equation}
as the L1 norm of the aggregated attention-gradient tensor for layer $\ell$, quantifying the layer's impact on the prediction. These weights are subsequently normalized across layers. 

We additionally incorporate a depth-based prior
\begin{equation}\label{eq:depth_prior}
    s_\ell = \frac{\exp(\lambda_d(\ell + 1))}{\sum_{k=1}^{L} \exp(\lambda_d(k + 1))}
\end{equation}
where $\lambda_d$ is the temperature. This assigns higher weights to deeper layers to emphasize semantic representations. 

These two components are combined and normalized:
\begin{equation}\label{eq:alpha_def}
\alpha_\ell = \frac{g_\ell s_\ell}{\sum_{k=1}^{L} g_k s_k},
\end{equation}
yielding layer-level weights $\alpha_\ell$ that balance empirical gradient evidence with architectural priors. This formulation allows strong gradient signals to override the depth bias when layers show exceptional importance for the prediction.

\paragraph{Relevance propagation}
For each generated token, we initialize a running relevance matrix
\begin{equation}
\mathbf{R} \leftarrow \mathbf{I}_N,
\end{equation}
where $\mathbf{I}_N$ is the identity matrix ensuring that every token initially 
contributes only to itself. We then propagate relevance through layers sequentially.
At layer $\ell$, we obtain the gradient-fused, row-normalized attention 
matrix $E_\ell$ (Eq.~\ref{eq:fused_attention}) and construct 
a layer-specific relevance transformation:

\begin{equation}
\mathbf{L}_\ell = \mathbf{I}_N + \alpha_\ell E_\ell,
\end{equation}

where $\alpha_\ell$ is the adaptive layer weight from Eq.~\eqref{eq:alpha_def}. Rather than computing the full matrix product across all layers~\cite{abnar2020attention}, which is prone to numerical instabilities and noise buildup, we employ additive accumulation, as in \cite{chefer2021generic}:

\begin{equation}\label{eq:additive_rollout}
\mathbf{R} \leftarrow \mathbf{R} + \mathbf{L}_\ell \mathbf{R}.
\end{equation}

With all modalities encoded in a single sequence $x$ of length $N$ (Eq.~\ref{eq:sequence_repr}), final relevancy matrix $\mathbf{R} \in \mathbb{R}^{N \times N}$ captures unified cross-modal interactions:
each row $\mathbf{R}_{t,:}$ ($t \in \{\mathcal{V}, \mathcal{P}, \mathcal{Y}\}$) scores elements from all modalities for relevance to the generation of token $t$.

\subsection{Cross-Modal Token Relevancy} 
\label{sec:token_relevancy}
To prevent informational evidence from being diluted by less meaningful tokens or even hallucinated detours, 
we introduce a cross-modal alignment weighting scheme that prioritizes tokens that are strongly 
associated with textual and visual input, and generated with high model confidence.

\paragraph{Prompt-Alignment Weight} For each generated token $t \in \mathcal{Y}$, we compute its alignment to the prompt by extracting relevance from the propagated matrix:
\begin{equation}
a_t = \frac{1}{|\mathcal{P}|} \sum_{i \in \mathcal{P}} \mathbf{R}(t,i)
\end{equation}
where $\mathcal{P}$ denotes prompt token indices and $\mathbf{R}(t,i)$ measures how strongly token $t$ addresses the prompt content semantically or referentially. 

\paragraph{Visual-Alignment Weight} Similarly, for prompt saliency computation, we define the visual-alignment weight:
\begin{equation}
v_t = \frac{1}{|\mathcal{V}|} \sum_{i \in \mathcal{V}} \mathbf{R}(t,i)
\end{equation}
where $\mathcal{V}$ denotes visual token indices and $v_t$ quantifies token $t$'s grounding in visual evidence supported generation. 

\paragraph{Confidence Weight} We define the model's confidence in token $t$ as its softmax probability:
\begin{equation}
p_t = \frac{\exp(z_t)}{\sum_{w \in \Omega} \exp(z_w)}
\end{equation}
where $z_t$ is the logit for token $t$ and $\Omega$ is the vocabulary. 
This reflects the model's certainty given the full context, capturing the degree of support for grounded predictions.

\paragraph{Combined Weighting} We define the alignment weight as
\begin{equation}
w_t^{(m)} = \begin{cases}
a_t & \text{for visual saliency } (m = \mathcal{V}) \\
v_t & \text{for prompt saliency } (m = \mathcal{P})
\end{cases}
\end{equation}
where $m \in \{\mathcal{V}, \mathcal{P}\}$ specifies the target modality. 
The final token weight integrates both confidence and alignment:
\begin{equation}\label{eq:combined_weighting}
\beta_t^{(m)} = \frac{p_t \cdot w_t^{(m)}}{\sum_{k \in \mathcal{Y}} p_k \cdot w_k^{(m)}}
\end{equation}
Thus, a token’s contribution $\beta_t^{(m)}$ to the saliency map of modality $m$ is determined by its alignment with the complementary modality, 
modulated by predictive confidence\nobreak\ $p_t$. 
\paragraph{Joint Token Relevance} To capture tokens' cross-modal relevance, we define joint token relevance:
\begin{equation}\label{eq:joint_relevance}
\gamma_t = \sqrt{\beta_t^{(\mathcal{V})} \times \beta_t^{(\mathcal{P})}}
\end{equation}
which identifies tokens exhibiting both strong prompt alignment and visual grounding, thereby capturing the interaction of multimodal reasoning within the generated response.

\paragraph{Relevance Flow Redistribution} Although function words (e.g., “is,” “of”) often carry high relevance in autoregressive prediction, 
they contribute minimally in semantic interpretation. To enhance interpretability, we optionally transfer relevance mass from each function word 
onto its syntactically linked content word (e.g., “is a bird”),  thereby sharpening explanatory emphasis on semantically  substantive elements. 

We define the normalized influence (left) and flow (right) across all token pairs (for $j>i$) as:
\begin{equation}
F_{i\to j} = \frac{\mathbf{R}(j,i)}{\sum_{k>i}\mathbf{R}(k,i)}; \quad f_{i\to j} = \beta_i^{(m)} \times F_{i\to j}
\end{equation}
where $F_{i\to j}$ captures the normalized connection strength between tokens and $f_{i\to j}$ represents the actual relevance flow, with $\sum_{j>i}F_{i\to j}=1$ conserving token $i$'s influence budget.
We update token weights by incorporating received flows from all preceding tokens:
\begin{equation}
\beta_t^{(m)'}  = \beta_t^{(m)} + \lambda_f \sum_{i<t}f_{i\to t}
\end{equation}
where $\lambda_f \in [0,1]$ controls flow strength, followed by L1 normalization. We then compute the redistributed token relevance using Eq.~\eqref{eq:joint_relevance} with the updated weights $\beta_t^{(m)'}$. 

This redistribution flow is intended only to enhance token relevance interpretability and is deliberately omitted from the holistic aggregation  (Sec.~\ref{sec:holistic_aggregation}), as function words carry decisive importance in autoregressive predictions and often produce clean and meaningful attribution maps.

\subsection{Holistic Saliency Aggregation}
\label{sec:holistic_aggregation} 
The holistic relevance map is aggregated from individual token maps using token weights obtained from Eq.~\eqref{eq:combined_weighting}:
\begin{equation}
\tilde{\mathbf{R}}_m = \sum_{t \in \mathcal{Y}} \beta_t^{(m)} \mathbf{R}(t,m)
\end{equation}
where $\mathbf{R}(t,m)$ denotes the relevance vector from token $t$ to target modality $m$.

This produces modality-specific relevance vectors $\tilde{\mathbf{R}}_\mathcal{V}$ and $\tilde{\mathbf{R}}_\mathcal{P}$ 
that encode the joint contributions of image patches, prompt context, and the model's visual and textual reasoning. 
These holistic cross-modal saliency maps provide complementary explanatory views:

\begin{enumerate}
    \item \textbf{Spatial heatmap:} $\tilde{\mathbf{R}}_\mathcal{V}$ projects per-patch, prompt-conditioned relevance onto the original image, revealing the visual regions most critical in addressing the prompt and generating the output.
    \item \textbf{Prompt-saliency map:} $\tilde{\mathbf{R}}_\mathcal{P}$ quantifies the visual-conditioned contribution of individual prompt tokens in directing the model's focus to specific image areas that underlie the generated response.
    \item \textbf{Token relevance:} $\gamma_t$ captures the cross-modal relevance of each generated token, identifying words that exhibit both strong prompt alignment and visual grounding.
\end{enumerate}

\begin{figure}[ht]
\centering
\begin{subfigure}[b]{0.48\textwidth}
    \centering
    \scriptsize\textbf{(a) Q:} \textcolor{black}{Which} \textbf{\textcolor{red!80}{screen}} \textcolor{black}{looks} \textbf{\textcolor{red!50}{better?}} \\
    \textbf{A:} \textcolor{black}{The} \textcolor{orange!70}{left} \textcolor{black}{screen} \textcolor{black}{appears} \textcolor{black}{to} \textcolor{black}{have} \textcolor{black}{better} \textcolor{black}{visibility} \textcolor{black}{and} \textcolor{black}{clarity,} \textcolor{black}{as} \textcolor{black}{it} \textcolor{black}{displays} \textcolor{black}{a} \textbf{\textcolor{orange!89}{vibrant}} \textbf{\textcolor{orange!100}{cityscape}} \textcolor{black}{with} \textbf{\textcolor{orange!92}{distinct}} \textcolor{orange!58}{details,} \textcolor{black}{while} \textcolor{black}{the} \textcolor{orange!70}{right} \textcolor{black}{shows} \textcolor{orange!55}{dimmer} \textbf{\textcolor{orange!91}{image}} \textcolor{orange!83}{of} \textbf{\textcolor{orange!85}{desert}} \textcolor{orange!50}{landscape} \textcolor{black}{that} \textcolor{black}{is} \textcolor{black}{less} \textcolor{orange!50}{vivid}.
    \vspace{0.1cm}\\
    \includegraphics[width=\textwidth]{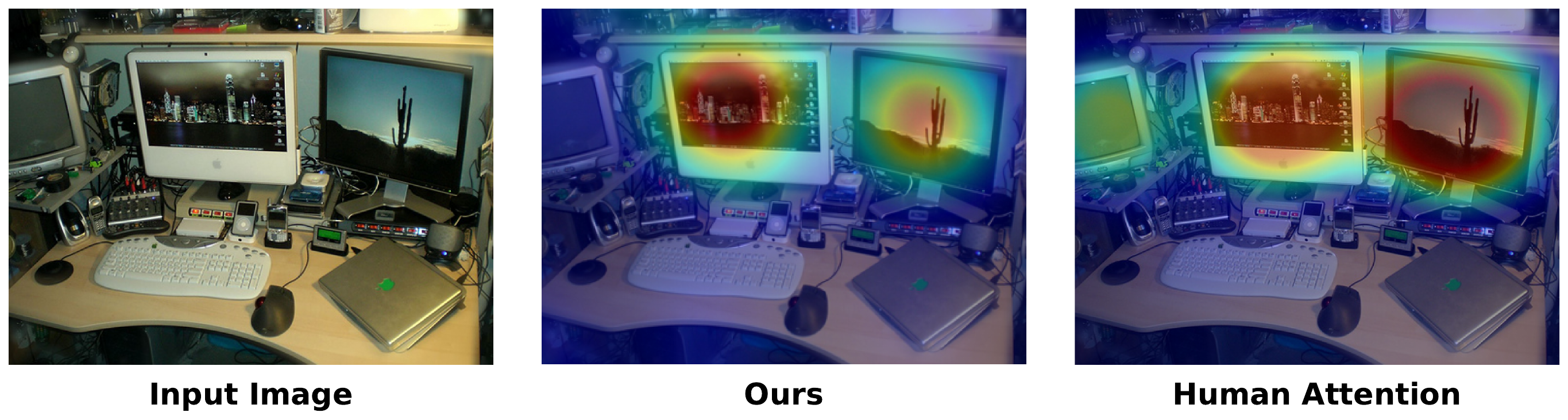}
    \vspace{0.03cm}
\end{subfigure}
\begin{subfigure}[b]{0.48\textwidth}
    \centering
    \scriptsize\textbf{(b) Q:} \textcolor{black}{What} \textcolor{black}{type} \textcolor{black}{of} \textbf{\textcolor{red!100}{condiment}} \textcolor{red!59}{is} \textcolor{black}{on} \textcolor{black}{the} \textcolor{black}{top} \textcolor{black}{shelf} \textcolor{red!50}{second} \textcolor{black}{from} \textcolor{black}{the} \textbf{\textcolor{red!65}{right?}} \\
    \textbf{A:} \textcolor{black}{The} \textbf{\textcolor{orange!70}{condiment}} \textcolor{black}{on} \textcolor{black}{the} \textcolor{orange!70}{top} \textcolor{black}{shelf} \textcolor{orange!50}{second} \textcolor{black}{from} \textcolor{orange!60}{right} \textbf{\textcolor{orange!100}{is}} {\textcolor{orange!50}{a}} \textcolor{orange!60}{jar} \textbf{\textcolor{orange!80}{of}} \textbf{\textcolor{orange!100}{mustard.}}
    \vspace{0.1cm}\\
    \includegraphics[width=\textwidth]{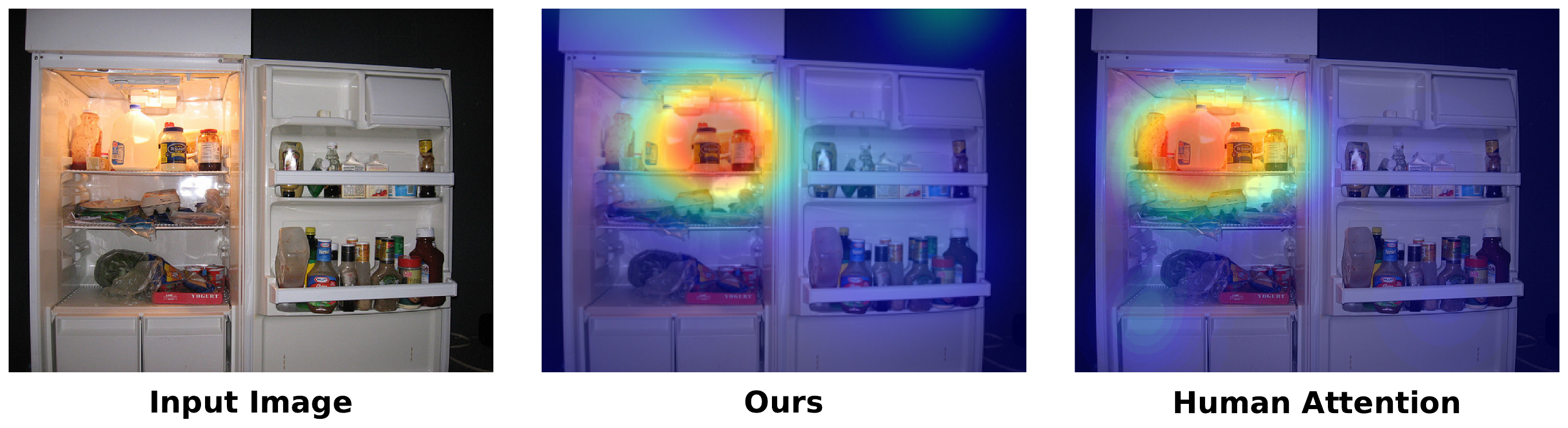}
\end{subfigure}
\vspace{-0.5cm}
\caption{\textbf{GLIMPSE saliency maps.} \textcolor{red}{Question tokens} are colored proportional to prompt saliency $\tilde{\mathbf{R}}_\mathcal{P}$;
\textcolor{orange}{response tokens} are colored proportional to cross-modal relevance $\gamma_t$; 
the heatmap intensity reflects the model's visual saliency $\tilde{\mathbf{R}}_\mathcal{V}$ over image regions.}
\label{fig:glimpse_example}
\vspace{-0.2cm}
\end{figure}

Taken together, these tripartite projections constitute a unified explanatory paradigm that elucidates 
  how salient image regions, the semantic influence of prompt queries, 
  and the relative informativeness of produced tokens converge to orchestrate the model's complete response generation. Figure~\ref{fig:glimpse_example} illustrates these capabilities.

\section{Experiments}
\label{sec:experiments}
We conduct two complementary experiments to evaluate GLIMPSE's interpretability capabilities. 
First, we assess the alignment between GLIMPSE's saliency maps and human attention (Sec.~\ref{subsec:alignment_experiment}), 
providing an objective benchmark for interpretability.
Second, we evaluate faithfulness through deletion and insertion experiments (Sec.~\ref{subsec:faithfulness_experiment}), 
measuring how well GLIMPSE's explanations reflect the model's actual decision-making process.

\subsection{Human Alignment Experiment}
\label{subsec:alignment_experiment}

We assess the alignment between GLIMPSE's saliency maps and human attention using the VQA-HAT~\cite{das2016attention} dataset, 
which provides fine-grained human generated heatmaps for VQA tasks. 
We restrict our evaluation to open-ended questions to align with the free-form generative setting addressed by GLIMPSE.
In consideration of intercoder reliability, we further subset the QA set using only samples with at least 3 annotator maps, which are then averaged into a single heatmap per sample. 
For all experiments, we use Qwen-VL 2.5 (32B)~\cite{qwen25vl2025} as our backbone LVLM.

\subsubsection{Evaluation Metrics}
We report two complementary alignment scores computed against the aggregated human attention maps:

\textbf{Normalized Scanpath Saliency (NSS)} – Mean normalized saliency at attention points:
\begin{equation}
\text{NSS} = \frac{1}{|B|} \sum_{(i,j) \in B} \frac{\tilde{R}_{i,j} - \mu_{\tilde{R}}}{\sigma_{\tilde{R}}}
\end{equation}
where $\tilde R$ denotes the model saliency map; $B$ is the set of human‐attention locations above the $\theta$th percentile (with $\theta=95$ to capture high‐intensity regions). And $\mu_{\tilde R},\sigma_{\tilde R}$ are its mean and standard deviation.

\textbf{Spearman Rank Correlation} – Rank-order correlation coefficient between model saliency and human attention.

\subsubsection{Baselines}
We evaluate against representative attention-based, gradient-based, propagation-based, and hybrid explainers. 
Each baseline is extended to produce sequence-level saliency maps as summarized in Table~\ref{tab:baselines}. 
We include both ``TMME (vanilla)'' and ``TMME (last 12$\ell$)'' variants, as we observed depth-dependent noise buildup in deep LVLMs causes vanilla TMME to perform poorly, and we therefore introduce a last-12-layer variant for fairer comparison.

\begin{table}[t]
    \centering
    \footnotesize
    \setlength{\tabcolsep}{4pt}
    \begin{tabular}{p{1.4cm}p{2.4cm}p{3.6cm}}
        \toprule
        \textbf{Family} & \textbf{Method} & \textbf{Sequence-level Adaptation} \\
        \midrule
        Attention & Raw Attention & Raw attention averaged across layers; per-token maps averaged over sequence. \\
        \midrule
        Attention Propagation & Rollout~\cite{abnar2020attention} & Rollout applied to all layers; per-token maps averaged. \\
        \midrule
        Gradient & Grad-CAM~\cite{selvaraju2017gradcam} & Gradients w.r.t. final layer; per-token maps averaged. \\
        \midrule
        Hybrid & TMME (vanilla)~\cite{chefer2021generic} & Propagation applied to all layers; per-token maps averaged. \\
        \midrule
        Hybrid & TMME (last 12$\ell$) & Only the last 12 layers; per-token maps averaged. \\
        \bottomrule
    \end{tabular}
    \caption{Baseline explainers and their sequence-level adaptations.}
    \label{tab:baselines}
\end{table}

\subsubsection{Quantitative Results}
\label{subsubsec:quantitative_results}

Table~\ref{tab:quantitative_results} presents the quantitative comparison results, demonstrating GLIMPSE outperforms in alignment with human attention across all evaluation metrics, with a rank correlation of 0.250 and NSS of 1.014.

We observe a stark gap between vanilla TMME and its 12-layer variant: propagating relevance through all layers yields poor 
performance, whereas restricting propagation to the final layers recovers substantially better alignment by mitigating early-layer noise accumulation. 
In contrast, GLIMPSE employs full propagation and yet achieves substantial improvements over the partially propagated TMME variant, demonstrating the efficacy of its relevancy-based layer weights and depth-aware propagation.

\begin{table}[t]
    \centering
    \footnotesize
    \setlength{\tabcolsep}{4pt}
    \begin{tabular}{l c c}
        \toprule
        \textbf{Method} & \textbf{NSS} $\uparrow$ & \textbf{Rank Correlation} $\uparrow$ \\
        \midrule
        Raw Attention & 0.485 $\pm$ 0.033 & 0.015 $\pm$ 0.009 \\
        Attention Rollout & -0.082 $\pm$ 0.016 & -0.010 $\pm$ 0.009 \\
        Grad-CAM & 0.267 $\pm$ 0.025 & 0.020 $\pm$ 0.008 \\
        TMME (vanilla) & -0.205 $\pm$ 0.013 & -0.153 $\pm$ 0.011 \\
        TMME (last 12$\ell$) & 0.591 $\pm$ 0.031 & 0.171 $\pm$ 0.010 \\
        \midrule
        \textbf{GLIMPSE (ours)} & \textbf{1.014 $\pm$ 0.032} & \textbf{0.250 $\pm$ 0.008} \\
        \bottomrule
    \end{tabular}
    \caption{\textbf{Human alignment experiment results.}
    GLIMPSE demonstrates superior performance across all metrics, 
    with improvements of +71.5\% in NSS and +46.2\% in rank correlation over TMME last 12$\ell$.}
    \label{tab:quantitative_results}
    \vspace{-0.2cm}
\end{table}

\subsection{Faithfulness Experiment}
\label{subsec:faithfulness_experiment}
While high alignment scores indicate 
that a model's explanations correspond to human attention, they do not necessarily guarantee faithful attribution of the model's internal reasoning.
In addition, alignment provides an objective benchmark for interpretability, yet it can over-penalize an explainer that (i) discovers alternative yet valid visual cues or (ii) 
follows the model's hallucinations instead of true diagnostic regions  (see Sec.~\ref{sec:qualitative}). Consequently, human-alignment metrics may underestimate GLIMPSE's comprehensive explanatory capabilities.

To measure faithfulness, we perform deletion and insertion experiments. Deletion removes the top-ranked image patches in descending saliency order, whereas insertion restores the same tokens onto an initially blurred image. The model's mean self-log-likelihood is linearly normalized for better comparison using the blurred baseline and the unperturbed response. 
Integrating this confidence curve and dividing by the perturbation span yields a normalized area under the curve (AUC). We report normalized AUCs at top 5\%, 15\%, and 30\% patch perturbation as these early-stage intervals isolate the influence of the most salient regions, where different explainers diverge most. A lower AUC for deletion (rapid confidence collapse) and a higher AUC for insertion (rapid confidence recovery) indicate that the highlighted regions exert stronger causal influence on the model's prediction. 

Further, we also evaluate against human attention maps as a pseudo-ground-truth. While human attention is not a perfect oracle, they approximate the behavior of an ideal explainer and provides a valuable gold-standard reference.
\subsubsection{Quantitative Results}
\label{subsubsec:quantitative_results_faithfulness}
Table~\ref{tab:faithfulness_results} presents the evaluation results. GLIMPSE demonstrates strong performance across all perturbation levels, achieving the best AUC scores across baselines. 
Most baseline explainers yield negative or near-zero insertion AUC at top 5\% , indicating that their computed highest salient patches actually reduced model confidence below that of a blank image. 
By contrast, GLIMPSE attains a positive insertion AUC that closely matches the human‐attention, and its deletion AUCs remain on par with human-attention across all perturbation levels, indicating that it pinpoints important diagnostic evidence consistently.

\begin{table}[t]
    \centering
    \footnotesize
    \setlength{\tabcolsep}{3pt}
    \begin{tabular}{l c c c c c c}
        \toprule
        \multirow{2}{*}{\textbf{Method}} & \multicolumn{3}{c}{\textbf{Deletion AUC} $\downarrow$} & \multicolumn{3}{c}{\textbf{Insertion AUC} $\uparrow$} \\
        \cmidrule(lr){2-4} \cmidrule(lr){5-7}
        & \textbf{5\%} & \textbf{15\%} & \textbf{30\%} & \textbf{5\%} & \textbf{15\%} & \textbf{30\%} \\
        \midrule
        Raw Attention & 0.904 & 0.813 & 0.723 & 0.081 & 0.115 & 0.249 \\
        Attention Rollout & 0.975 & 0.945 & 0.901 & -0.089 & 0.032 & 0.133 \\
        Grad-CAM & 0.945 & 0.872 & 0.782 & 0.027 & 0.124 & 0.251 \\
        TMME (vanilla) & 0.979 & 0.943 & 0.896 & 0.078 & 0.120 & 0.194 \\
        TMME (last 12$\ell$) & 0.937 & 0.802 & 0.719 & -0.005 & 0.170 & 0.321 \\
        \midrule
        \textbf{GLIMPSE (ours)} & \textbf{0.855} & \textbf{0.718} & \textbf{0.617} & \textbf{0.134} & \textbf{0.276} & \textbf{0.424} \\
        \midrule
        Human Attention & 0.852 & 0.707 & 0.589 & 0.149 & 0.344 & 0.502 \\
        \bottomrule
    \end{tabular}
    \caption{\textbf{Faithfulness experiment results.} Deletion AUC: lower is better. Insertion AUC: higher is better.}
    \label{tab:faithfulness_results}
    \vspace{-0.4cm}
\end{table}

\subsection{Ablation Study}
\label{subsec:ablation}

To assess GLIMPSE's sensitivity to its design choices, we performed a comprehensive ablation study over key components: token saliency weighting, fusion strategy, layer weighting, and propagation depth. Table~\ref{tab:ablation} reports the mean NSS and rank correlation for each variant.

\textbf{Token Saliency Components.} Token confidence weighting has proven to be critical, with its removal causing a 21.3\% NSS drop. Dropping both token confidence and prompt weighting yields a greater drop in performance, indicating these components play complementary roles in modulating the individual token contribution.

\textbf{Layer Weighting.} Depth weighting is the most essential component among all factors, 
removing it causes performance to collapse to negative values (NSS=-0.210), demonstrating that without proper weighting, 
early-layer noise overwhelms meaningful signals.

\textbf{Propagation Depth.} Dropping propagation to fewer layers steadily degrades performance, in contrast to what we observe with 
TMME~\cite{chefer2021generic} where subsetting final layers substantially boosts performance. 
This validates that our layer weighting strategy effectively mitigates early-layer noise while facilitating information flow. 
Notably, using last 30\% layers without depth weighting drastically reduces performance to NSS=0.490, 
underscoring that our depth weighting scheme does more than merely suppress low‐level noise. It effectively rescales and preserves informative signals from early‐layer features when warranted.

These findings underscore that well-designed weighting schemes constitute the cornerstone of robust interpretability in deep multimodal networks.

We report the hyperparameter configuration that achieves the best quantitative performance; 
in practice, although full propagation yields optimal scores, but using last 60\% of depth-weighted layers trades a marginal performance loss for efficiency gains.
Moreover, we observed that removing punctuation, and when the model permits, 
adding a brief system instruction cue to localize salient image regions before answering can encourage more concentrated heat-maps. 
Optionally, a light Gaussian blur can be applied for additional aesthetic refinement.

\begin{table}[t]
    \centering
    \footnotesize
    \setlength{\tabcolsep}{4pt}
    \begin{tabular}{l l c c}
        \toprule
        \textbf{Component} & \textbf{Setting} & \textbf{NSS} $\uparrow$ & \textbf{Rank Corr.} $\uparrow$ \\
        \midrule
        \multirow{4}{*}{Token Saliency} & \textbf{Full (baseline)} & \textbf{1.014} & \textbf{0.250} \\
                                       & w/o prompt weighting & 0.899 & 0.203 \\
                                       & w/o token confidence & 0.798 & 0.185 \\
                                       & w/o both & 0.780 & 0.182 \\
        \midrule
                \multirow{4}{*}{Fusion Strategy} & \textbf{Adaptive (temp=0.5)} & \textbf{1.014} & \textbf{0.250} \\
                                         & Simple average & 0.950 & 0.234 \\
                                         & Temperature = 0.2 & 1.012 & 0.248 \\
                                         & Temperature = 1.0 & 1.011 & 0.245 \\
        \midrule
                \multirow{6}{*}{Layer Weighting} & \textbf{Full (depth temp=0.2)} & \textbf{1.014} & \textbf{0.250} \\
                                         & w/o depth weighting & -0.210 & -0.167 \\
                                         & w/o layer relevance & 0.918 & 0.213 \\
                                         & Depth temp = 0.5 & 0.911 & 0.215 \\
                                         & Depth temp = 1.0 & 0.883 & 0.209 \\
        \midrule
        \multirow{4}{*}{Propagation Depth} & \textbf{All layers (baseline)} & \textbf{1.014} & \textbf{0.250} \\
                                          & Last 60\% (38 layers) & 1.011 & 0.247 \\
                                          & Last 30\% (20 layers) & 0.984 & 0.237 \\
                                          & 30\% w/o depth weight & 0.670 & 0.178 \\
        \bottomrule
    \end{tabular}
    \caption{\textbf{Comprehensive ablation study.}}
    \label{tab:ablation}
    \vspace{-0.2cm}
\end{table}

\section{Qualitative Analysis}
\label{sec:qualitative}
\subsection{Token Relevancy}
\label{subsec:token_relevancy}

Tokens are not created equal. Within an autoregressive LVLM, different lexical units contribute unequally to address the prompt and attend to distinct visual evidence. Because GLIMPSE computes a fully propagated relevance for each token, it exposes this heterogeneity: we can inspect how each generated token aligns with the prompt and which image regions it recruits, before those signals are fused into the holistic saliency map.

\begin{figure}[t]
    \centering
    \scriptsize
    \textbf{Q:} \textcolor{black}{What} \textcolor{black}{is} \textcolor{red!60}{in} \textcolor{black}{the} \textbf{\textcolor{red!90}{dishes?}}\\
    \textbf{A:} \textcolor{black}{The} \textcolor{black}{dishes} \textcolor{black}{contain} 
    \textcolor{black}{various} \textcolor{black}{foods:} \textcolor{black}{one} \textcolor{black}{has} 
    \textbf{\textcolor{orange!100}{rice}} \textcolor{black}{with} \textcolor{black}{a} \textbf{\textcolor{orange!80}{serving}} 
    \textcolor{orange!75}{spoon,} \textcolor{orange!50}{another} \textcolor{orange!80}{dark} 
    \textbf{\textcolor{orange!90}{sauce}} \textcolor{black}{or} \textcolor{orange!80}{soup} \textcolor{black}{with} 
    \textcolor{orange!50}{vegetables} \textcolor{black}{and} \textcolor{orange!55}{possibly} \textcolor{orange!75}{meat,} 
    \textcolor{black}{and} \textcolor{black}{there} \textcolor{black}{are} \textcolor{black}{several} 
    \textcolor{black}{plates} \textcolor{black}{of} \textcolor{black}{different} \textcolor{black}{cooked} 
    \textcolor{black}{dishes} \textcolor{black}{including} \textcolor{black}{what} \textcolor{black}{appears} 
    \textcolor{black}{to} \textcolor{black}{be} \textcolor{orange!50}{stir-fried} \textcolor{orange!60}{vegetables,} 
    \textcolor{orange!45}{noodles,} \textcolor{black}{and} \textcolor{orange!45}{other} \textcolor{orange!50}{Asian-style} 
    \textcolor{black}{meals.}
    \vspace{0.2cm}
    
    \begin{subfigure}[b]{0.48\textwidth}
        \centering
        \includegraphics[width=\textwidth]{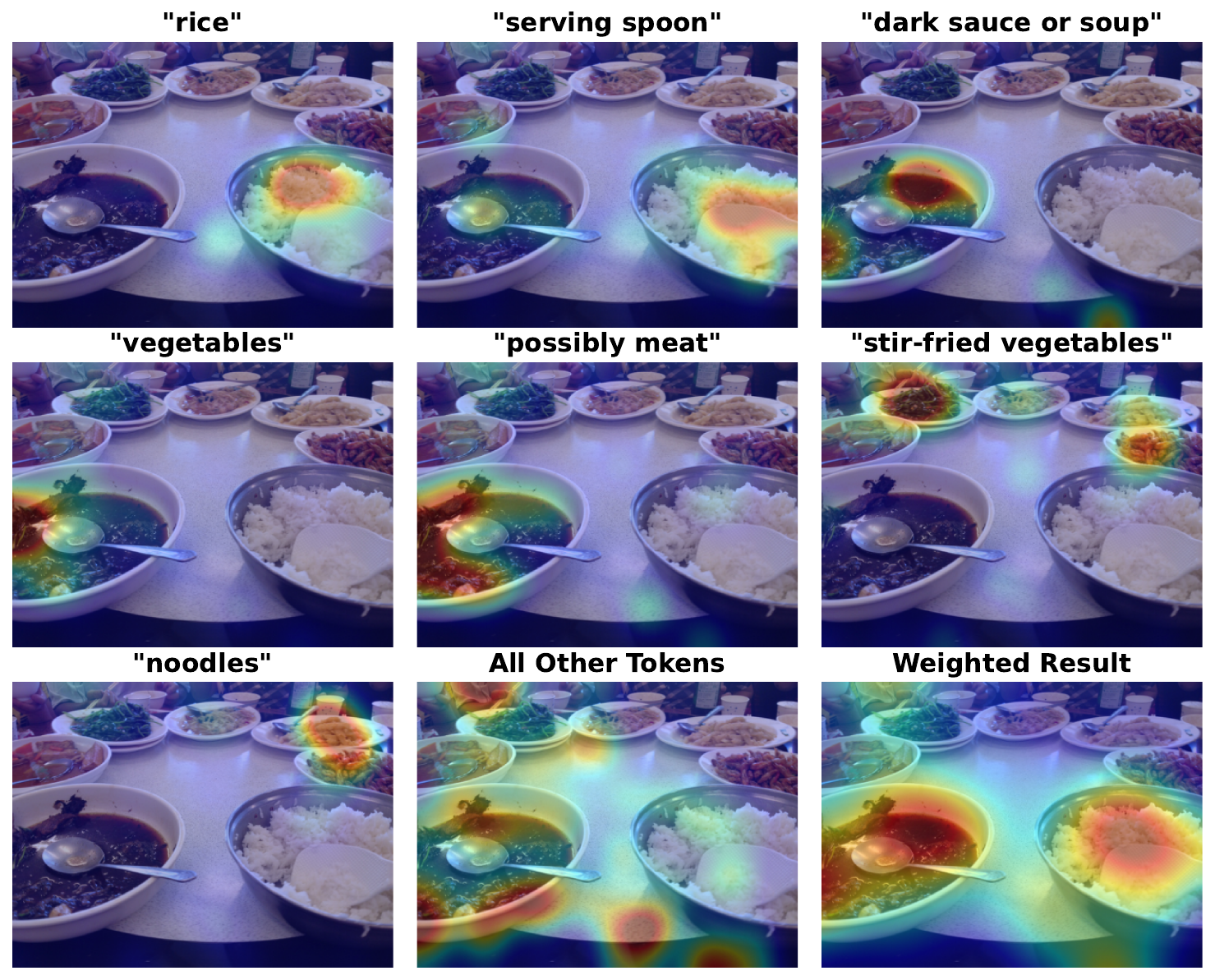}
        \caption{Saliency map for high-relevancy token groups, 
        revealing the specific image regions that contribute to each token's generation. 
        The final saliency map is aggregated from all token-level maps weighted by respective saliency scores}
        \label{fig:token_relevancy_heatmaps}
    \end{subfigure}
    \hfill
    \begin{subfigure}[b]{0.48\textwidth}
        \centering
        \includegraphics[width=\textwidth]{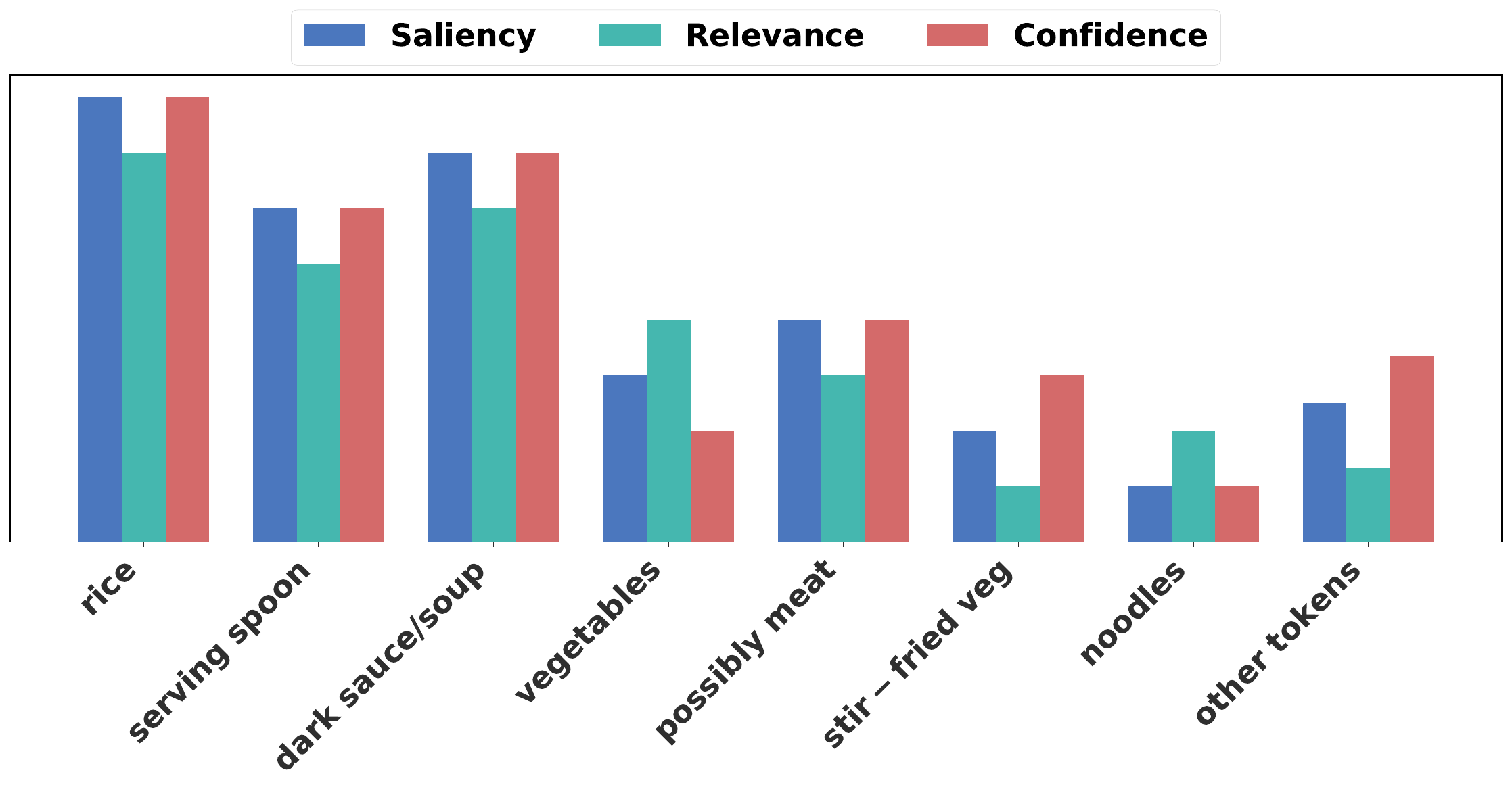}
        \caption{GLIMPSE-computed token saliency scores across the generated response.
        \textbf{Relevance} measures propagated gradient attention relevance; 
        \textbf{Confidence} represents softmax generation confidence.
        \textbf{Saliency} combines both relevance and confidence; }
        \label{fig:token_relevancy_chart}
    \end{subfigure}
    \caption{\textbf{Token-level relevancy.} We present spatial saliency maps (a) and saliency scores (b) for semantically meaningful token groups. 
    Stop-words and punctuation are excluded.}
    \label{fig:token_relevancy}
    \vspace{-0.3cm}
\end{figure}

Figure~\ref{fig:token_relevancy_heatmaps} demonstrates that GLIMPSE achieves accurate token-level localization, revealing distinct visual grounding for individual tokens in the generated response.

\textbf{Foreground-bias diagnosis.}
To answer the question ``What is in the dishes?'', the model first generates tokens such as rice, serving spoon, and dark sauce or soup, all of which refer to the largest and closest objects in the foreground. As shown in Figure~\ref{fig:token_relevancy_chart}, GLIMPSE assigns these tokens the highest prompt-relevance and confidence scores. Only afterward does the model mention more distant items (e.g., vegetables), which receive lower saliency. In this particular example, that high-to-low saliency progression corresponds closely with the spatial progressive reasoning exhibited by the model, and how it sequences its token generation, which mirrors the human tendency to emphasize prominent foreground objects over background elements. This provides clear evidence that GLIMPSE faithfully uncovers the model's internal stages of visual reasoning.

\textbf{Error localization and down-weighting.}
Token-level attribution also aids in identifying fractional discrepancies. In this example, the model's response is only partially correct: it hallucinates the presence of noodles. As shown in Figure~\ref{fig:token_relevancy_heatmaps}, the corresponding heatmap for that token illuminates a region that bears no noodle-like features---demonstrating that the model's attribution for ``noodles'' is unsupported by the actual visual content. GLIMPSE correctly assigns low saliency driven by low confidence. As a result, it carries smaller weight in the final saliency map. 

By modeling outputs as a spectrum of contributions, GLIMPSE highlights high-confidence, well-grounded tokens and suppresses weak signals, enabling precise, fine-grained, and interpretable diagnostics of model decisions.

\subsection{Human Alignment}
\label{subsec:human_alignment}
Prior work \cite{xing2025attention} applied the IGOS++ based explainer and evaluated it on the same human attention dataset, VQA-HAT. 
They reported negative rank correlation (-0.20), and concluded that LVLMs focus on regions that significantly diverge from human attention. 
However, their results were obtained with perturbation-based optimization, 
and the low alignment may be artifacts of an objective that rewards pixelated sparsity over holistic evidence.

We argue instead that LVLMs aren't intrinsically misaligned—they've simply been viewed through the wrong lens. 
By using an extraction method that (i) respects autoregressive generation and (ii) leverages cross-modal token dependencies, 
our method's alignment jumps to +0.25 rank correlation. 
Compared to the -0.20 rank correlation reported by \cite{xing2025attention}, these results highlight that attribution quality is highly sensitive to the choice of explanation method.
While our evaluation uses a different backbone and may involve a different subset of VQA-HAT, 
the findings nonetheless suggest that meaningful alignment is achievable with appropriately tailored techniques.

\subsection{Human Misalignment}
Despite improved alignment, we investigate cases where GLIMPSE saliency maps diverge from human attention, aiming to identify potential sources of misalignment and understand the nature of machine visual reasoning.
\begin{figure}[t]
    \centering
    \begin{subfigure}[b]{0.48\textwidth}
        \centering
        \scriptsize
        \textbf{(a) Q:} \textcolor{black}{What} \textcolor{red!50}{type} \textcolor{red!60}{of} \textbf{\textcolor{red!80}{court}} \textcolor{black}{is} \textcolor{black}{this?}\\
        \textbf{A:} \textcolor{black}{This} \textcolor{black}{is} \textcolor{black}{a} \textbf{\textcolor{orange!80}{clay}} \textcolor{orange!60}{tennis} \textcolor{black}{court.} \textcolor{black}{The} \textcolor{orange!60}{orange} \textcolor{black}{surface} \textcolor{black}{and} \textcolor{black}{the} \textcolor{black}{visible} \textbf{\textcolor{orange!80}{lines}} \textcolor{black}{are} \textcolor{black}{characteristic} \textcolor{black}{of} \textcolor{black}{clay} \textcolor{black}{courts} \textcolor{black}{commonly} \textcolor{black}{used} \textcolor{black}{in} \textcolor{black}{professional} \textcolor{black}{tennis} \textcolor{black}{tournaments.}
        \vspace{0.1cm}\\
        \includegraphics[width=\textwidth]{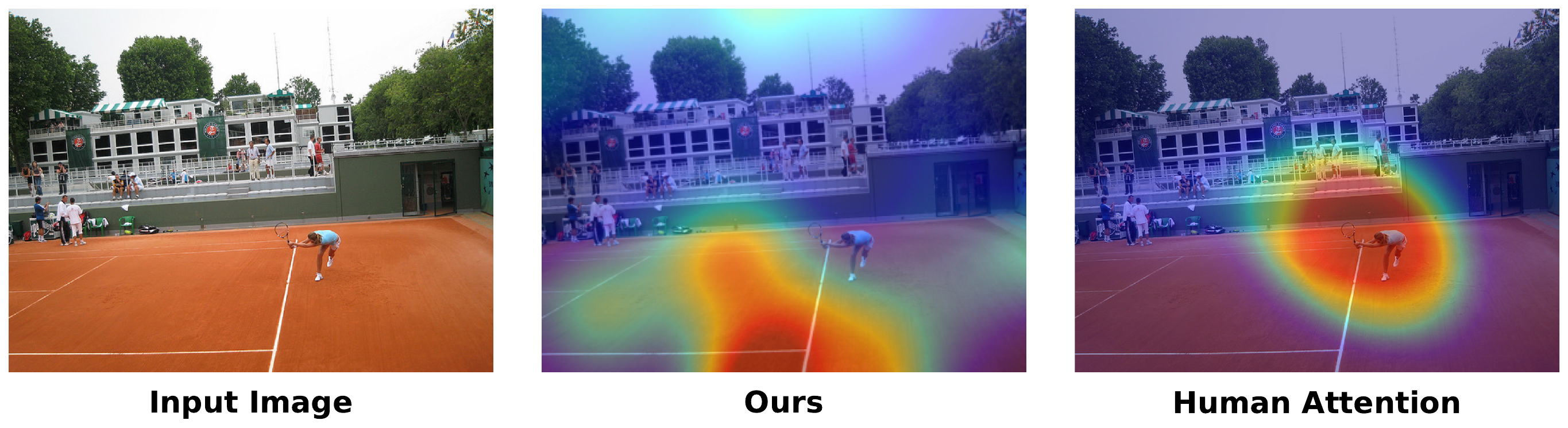}
        \label{fig:misalignment_case1}
    \end{subfigure}
    \hfill
    \begin{subfigure}[b]{0.48\textwidth}
        \centering
        \scriptsize
        \textbf{(b) Q:} \textcolor{black}{Are} \textcolor{black}{they} \textcolor{red!60}{going} \textbf{\textcolor{red!70}{forward}} \textcolor{black}{or} \textbf{\textcolor{red!70}{backward?}}\\
        \textbf{A:} \textcolor{black}{The} \textcolor{black}{people} \textcolor{black}{in} \textcolor{black}{the} \textcolor{orange!50}{boat} \textcolor{black}{appear} \textcolor{black}{to} \textcolor{orange!50}{be} \textcolor{orange!50}{rowing} \textcolor{black}{backward,} \textcolor{black}{as} \textcolor{black}{indicated} \textcolor{black}{by} \textcolor{black}{the} \textcolor{black}{direction} \textcolor{black}{of} \textcolor{black}{the} \textcolor{orange!60}{oars} \textcolor{black}{and} \textcolor{black}{the} \textbf{\textcolor{orange!90}{splash}} \textcolor{black}{created} \textcolor{black}{by} \textcolor{black}{the} \textcolor{black}{water.}
        \vspace{0.1cm}\\
        \includegraphics[width=\textwidth]{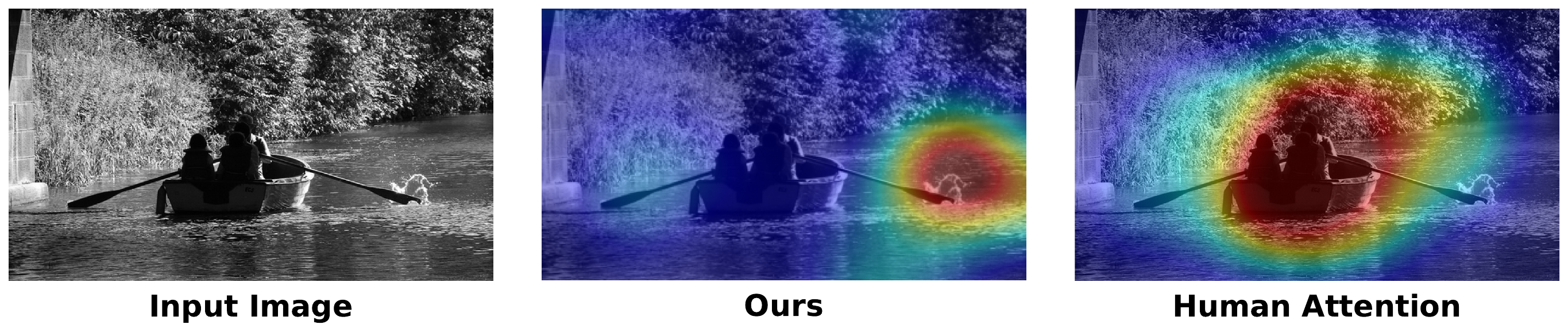}
        \label{fig:misalignment_case2}
    \end{subfigure}
    \vspace{-0.7cm}
    \caption{\textbf{Human attention misalignment examples.} (a) Humans focus on the player; GLIMPSE reveals the model's focus on the clay surface and court lines. (b) Human fixations track the boat body; GLIMPSE highlights the oars and water splash. Despite misalignment, both answers provided by the model are correct.}
    \label{fig:human_misalignment}
    \vspace{-0.2cm}
\end{figure}

As discussed in Section~\ref{sec:experiments}, human alignment does not guarantee faithfulness. 
Conversely, a lack of alignment does not automatically imply that the model's visual evidence is invalid. 
Figure~\ref{fig:human_misalignment} illustrates examples where divergence occurs. In both examples, the model produces 
the correct answer in spite of negative scores. GLIMPSE reveals that the model identifies alternative but equally 
diagnostic visual cues (e.g., court surface and lines as opposed to players) compared to human attention.
Such mismatches highlight that models can exploit complementary but sufficient visual evidence,
echoing observations in structured-reasoning settings~\cite{li2022causalvidqa,zhang2024tv,selvaraju2019hint}. 
Understanding such behavior is essential for improving explanation methods, building trustworthy models, and developing alignment-oriented applications.
\subsection{Hallucination and Bias Analysis}
\label{subsec:hallucination_analysis}

We examine how GLIMPSE can help diagnose hallucination and bias in generative LVLM outputs. Understanding why vision--language models make incorrect inferences requires more than measuring output accuracy, it 
demands an attribution-based examination of the visual cues that drive a model's reasoning. 
\begin{figure}[t]
    \centering
    \begin{subfigure}[b]{0.48\textwidth}
        \centering
        \scriptsize
        \textbf{(a) Q:} \textcolor{black}{What} \textbf{\textcolor{red!80}{sex}} \textcolor{red!60}{uses} \textcolor{black}{this} \textbf{\textcolor{red!90}{restroom?}}\\
        \textbf{A:} \textcolor{black}{This} \textcolor{black}{restroom} \textcolor{orange!60}{is} \textcolor{black}{designed} \textcolor{orange!80}{for} \textcolor{black}{use} \textcolor{black}{by} \textbf{\textcolor{orange!80}{both}} \textcolor{black}{males} \textcolor{black}{and} \textcolor{black}{females,} \textcolor{black}{as} \textcolor{black}{it} \textcolor{black}{includes} \textcolor{black}{urinals} \textcolor{black}{and} \textbf{\textcolor{orange!90}{sinks.}} \textcolor{black}{It} \textcolor{black}{appears} \textcolor{black}{to} \textcolor{black}{be} \textcolor{orange!50}{a} \textcolor{orange!50}{unisex} \textcolor{orange!70}{or} \textcolor{black}{gender-neutral} \textcolor{black}{facility.}\\
        \textbf{GT:} male
        \vspace{0.1cm}\\
        \includegraphics[width=\textwidth]{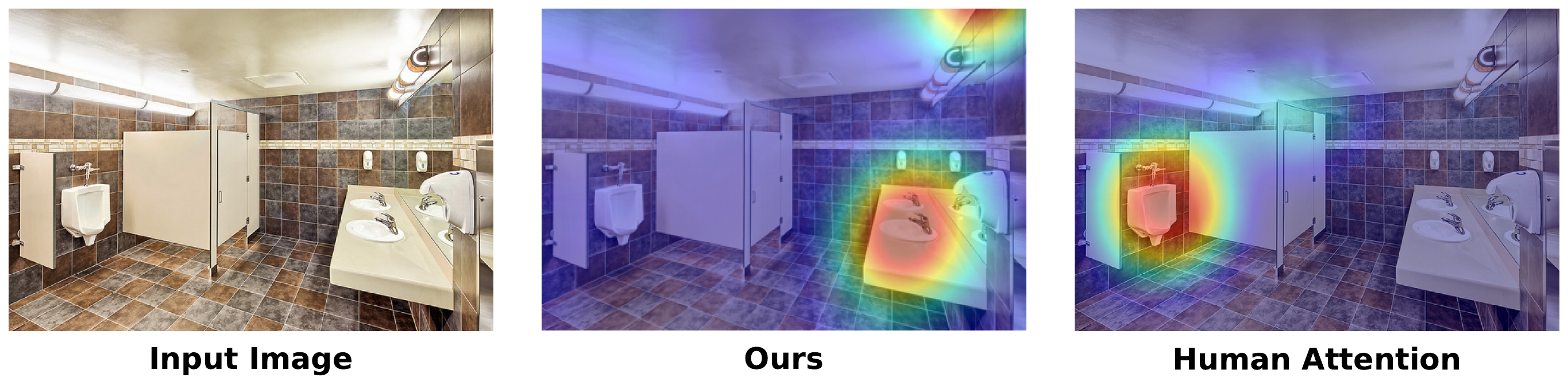}
        \label{fig:hallucination_case1}
    \end{subfigure}
    \hfill
    \begin{subfigure}[b]{0.48\textwidth}
        \centering
        \scriptsize
        \textbf{(b) Q:} \textcolor{black}{What} \textcolor{black}{kind} \textcolor{black}{of} \textbf{\textcolor{red!100}{light}} \textbf{\textcolor{red!96}{source}} \textcolor{black}{is} \textcolor{red!53}{in} \textcolor{black}{this} \textcolor{red!60}{room?}\\
        \textbf{A:} \textcolor{black}{The} \textcolor{black}{room} \textcolor{black}{has} \textcolor{orange!50}{a} \textbf{\textcolor{orange!80}{ceiling}} \textbf{\textcolor{orange!90}{fan}} \textcolor{orange!50}{with} \textcolor{black}{integrated} \textcolor{orange!60}{light} \textcolor{black}{fixture} \textcolor{black}{as} \textcolor{black}{the} \textcolor{black}{primary} \textbf{\textcolor{orange!80}{light}} \textcolor{black}{source.} \\
        \textbf{GT:} sun
        \vspace{0.1cm}\\
        \includegraphics[width=\textwidth]{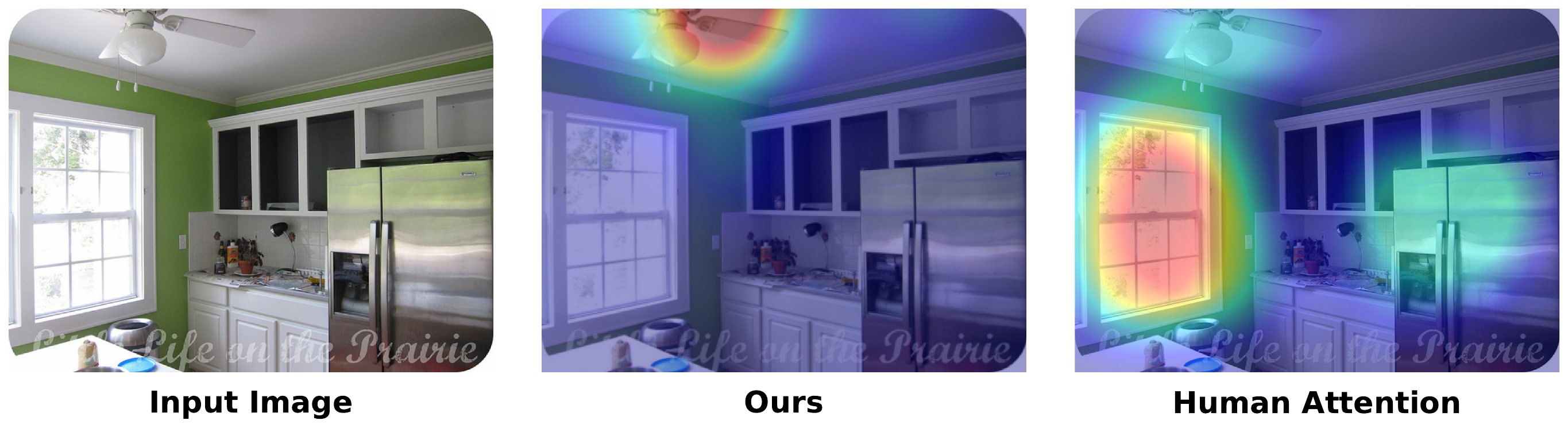}
        \label{fig:hallucination_case2}
    \end{subfigure}
    \vspace{-1cm}
    \caption{\textbf{Hallucination diagnosis examples.} (a) GLIMPSE reveals the model focuses on sinks rather than the male-only urinal, leading to a ``unisex'' 
    prediction (ground truth: male). (b) GLIMPSE shows the model's attention on the ceiling fan light while missing the sunlit window (ground truth: sun).}
    \label{fig:hallucination_analysis}
    \vspace{-0.2cm}
\end{figure}

Figure~\ref{fig:hallucination_analysis} illustrates two failure modes uncovered by GLIMPSE. 
When a model's prediction diverges from reality, we can trace that mistake back to 
misplaced saliency. In example (a), GLIMPSE reveals a bias toward ``default'' objects---sinks and counters---that 
outweighs the distinctive male-only feature, exposing a systematic under-weighting of less common but semantically 
critical elements. In example (b), the model's bias toward the most salient object (the ceiling fan light fixture) 
overrides a correct assessment of lighting intensity---brighter sunlight from the window, yet receives almost 
no attention.

This level of analysis enables a deeper, hypothesis-driven investigation of hallucination and bias. 
Rather than treating hallucinations as black-box anomalies, researchers can identify the exact visual 
evidence that misled the model, assess whether those patterns reflect dataset imbalances or architectural 
blind spots, and design targeted interventions (e.g., bias-aware fine-tuning, attention regularization, augmented 
supervision, or prompt engineering) to improve both faithfulness and fairness. In this way, attribution interpretation becomes 
a powerful tool for diagnosing and ultimately mitigating hallucinations in LVLMs.

\section{Conclusion}
We have shown that GLIMPSE achieves state-of-the-art alignment with human attention in explaining LVLM attribution, 
consistently outperforming prior methods in faithfulness while producing interpretable saliency maps. Looking ahead, 
we plan to extend GLIMPSE beyond static images into temporal settings such as video question answering. We hope this work contribute to advancing transparent, 
trustworthy AI systems, empowering researchers to diagnose failures, deepen understanding of model behavior, 
refine system design, and build models with better human alignment. 

{
    \small
    \bibliographystyle{ieeenat_fullname}
    \bibliography{main}
}

\end{document}